\documentclass{article}
\pdfoutput=1    

\usepackage{subcaption}
\usepackage[pdftex]{graphicx}
\DeclareGraphicsExtensions{.png,.pdf}
\usepackage{amssymb}
\usepackage{amsmath}
\usepackage{amsthm}

\usepackage[preprint]{nips_2018} 

\usepackage[utf8]{inputenc} 
\usepackage[T1]{fontenc}    
\usepackage{hyperref}       
\usepackage{url}            
\usepackage{booktabs}       
\usepackage{amsfonts}       
\usepackage{nicefrac}       
\usepackage{microtype}      

\title{Scalable Neural Network Compression and Pruning Using Hard Clustering and L1 Regularization}

\author{
  Yibo Yang, Nicholas Ruozzi, Vibhav Gogate \\
  Department of Computer Science \\
  University of Texas at Dallas \\
  Richardson, TX 75080 \\
  \texttt{\{yibo.yang, nicholas.ruozzi, vibhav.gogate\}@utdallas.edu} \\
}

\begin{document}

\maketitle

\begin{abstract}
We propose a simple and easy to implement neural network compression algorithm that achieves results competitive with more complicated state-of-the-art methods. The key idea is to modify the original optimization problem by adding $K$ independent Gaussian priors (corresponding to the $k$-means objective) over the network parameters to achieve parameter quantization, as well as an $\ell_1$ penalty to achieve pruning. Unlike many existing quantization-based methods, our method uses \textit{hard} clustering assignments of network parameters, which adds minimal change or overhead to standard network training. We also demonstrate experimentally that tying neural network parameters provides less gain in generalization performance than changing network architecture and connectivity patterns entirely. 

\end{abstract}

\section{Introduction}

Neural networks represent a family of highly flexible and scalable models that have rapidly achieved state-of-the-art performance in diverse domains such as computer vision \citep{krizhevsky2012, girshick2014, he2016} and speech \citep{hinton2012,deng2013}. However, the storage requirements of large, modern neural networks can make them impractical for applications with storage limitations (e.g., mobile devices). Moreover, as they are often trained on small datasets compared to their number of parameters, they can potentially overfit. \citet{denil2013} showed that a large proportion of neural network parameters are in fact not required for generalization performance, and interest in model compression has surged.

A variety of compression methods have been proposed including pruning \citep{lecun1990, han2015}, quantization \citep{han2016,ullrich2017,chen2015}, low-rank approximation \citep{denil2013, denton2014,jaderberg2014}, group lasso \citep{wen2016}, variational dropout \citep{molchanov2017}, 
etc.  Here, we focus on the quantization/parameter tying approach to compression combined with pruning.  

Growing literature has focused on automatic parameter tying, i.e., automatically discovering which parameters of the model should be tied together.  \citet{nowlan1992} proposed a soft parameter tying scheme based on a mixtures of Gaussians prior and suggested a gradient descent method to jointly optimize both the parameters of the network and the mixture model.  \citet{chen2015} proposed a random parameter tying scheme based on hashing functions. \citet{han2016} proposed a compression pipeline that involved thresholding to prune low-magnitude parameters, $k$-means clustering to tie parameters layer-wise, and a final retraining stage to fine-tune tied parameters.  This work demonstrated that high compression rates are achievable without much loss in accuracy.  Building on the work of \citep{nowlan1992}, \citet{ullrich2017} imposed a Gaussian mixture prior on network parameters to encourage clustering. At convergence, they proposed quantizing the parameters by assigning them to the mixture component that generates each parameter with highest probability.   \citet{louizos2017} proposed a full Bayesian approach to compression using scale mixture priors.  This approach has the advantage that posterior distributions can be used to estimate the significance of individual bits in the learned weights.  \citet{louizos2017} demonstrated that this approach can yield state-of-the-art compression results for some problems. \citet{agustsson2017} recently proposed a soft-to-hard quantization approach in which scalar quantization is gradually learned through annealing a softened version of quantization distortion; compression is achieved with low-entropy parameter distribution instead of pruning. 
Parameter tying via quantization has also been used in the graphical models community to scale up inference \cite{st-aubin&al00,gogate&domingos11a} and more recently to regularize and improve the prediction quality of parameter learning algorithms \cite{chou&al16a,chou&al18}.


While much previous work has demonstrated that significant compression can be achieved while preserving the accuracy of the final network (in many cases $\approx 1\%$ loss in accuracy), many of these approaches have potential drawbacks that can limit their application. The Gaussian mixture approach of \citet{nowlan1992} and \citet{ullrich2017} can be computationally expensive, as the time and memory requirements for backpropagation is increased $K$-fold under a $K$-component GMM prior, in addition to its large number of sensitive hyperparameters that can require extensive tuning. 
Moreover, the GMM objective itself suffers from well known (and often pathological) local minima issues.
The approach of \citet{han2016} uses separate pruning and parameter tying stages, which potentially limits its compression efficiency;  additionally, the required layer-wise codebook storage can become expensive for deep networks. 
The soft-to-hard quantization approach of \citep{agustsson2017} uses soft-assignment probabilities for network parameters like in the GMM approach, and gradually obtains hard assignment by annealing; by contrast, our method uses hard-assignment throughout and can therefore require much less computation. 
The full Bayesian approach, similar to the GMM approach, has a number of additional parameters to tune (e.g., constraints on variances, initialization of the variational parameters, etc.). The Bayesian approach also requires sampling for prediction (which can be done deterministically but with some additional loss).
In this paper, we show that such sophisticated methods may not be necessary to achieve good compression in practice.

This work tackles compression by quantization and sparsity inducing priors. For quantization, we consider an independent Gaussian prior, i.e., each parameter is non-probabilistically assigned to one of $K$ independent Gaussian distributions, and the prior penalizes each parameter by its $\ell_2$ distance to the mean of its respective Gaussian.  This prior places no restriction on which parameters can be tied together (e.g., parameters from the input could be tied to parameters into the output), reduces the number of hyperparameters that need to be tuned compared to standard Gaussian mixtures, and requires a small change to the typical gradient descent using only linear time and memory overhead. We observe that quantization alone is insufficient for the desired compression level, and introduce sparsity by adding a standard $\ell_1$ penalty on top of the quantization prior; we demonstrate experimentally that the combined prior yields state-of-the-art compression results. 


\section{Quantization by Parameter Tying}
We consider the problem of learning a neural network by minimizing the regularized loss function
$$\mathcal{L}(\boldsymbol{W}) = E_D(\boldsymbol{W}) + \lambda R(\boldsymbol{W}),$$
where $\boldsymbol{W}$ is the set of network parameters of size $N$, $E_D$ is the loss on training data $D$, and $R$ is a function chosen to induce desired properties in learned parameters, e.g., better generalization performance, which cannot be achieved by optimizing $E_D$ alone. $R$ is often chosen to be the $\ell_1$ or $\ell_2$ norm, which encourages sparse parameter vectors or bounded parameter vectors, respectively.

In this work, we achieve quantization with an alternative form of regularization. In a parameter-tied model, $\boldsymbol W$ is partitioned into $K$ sets, and parameters in each set are constrained to be equal, i.e., $\boldsymbol W$ contains only $K$ distinct values. Formally, let $\mathbf{C}=\{C_k\subseteq \{1,\ldots,N\}|k=1,...,K\}$ be disjoint sets, or clusters, of parameter indices, such that $\cup_{k=1}^K C_k = \{1,\ldots,N\}$. If the parameters indexed by $C_k$ are required to share the same value, then learning under parameter tying yields the constrained optimization problem of minimizing $\mathcal{L}(\boldsymbol W)$ subject to $  w_i = w_j, \forall k\in\{1,\ldots,K\}, i, j\in C_k$.


In the neural networks community, parameter tying is fundamental to convolutional neural networks (CNNs), where parameters of local receptive fields are shared across a specific filter. In practice, high-dimensional data sets may possess neither obvious structure nor prior information about how model parameters should be tied.  This motivates our goal of discovering which parameters should be tied without prior knowledge, i.e., automatic parameter tying, in which we optimize with respect to both the parameters and the cluster assignments. In general, this problem will be intractable as the number of possible partitions of the parameters into clusters, the Bell number, grows exponentially.

Instead, we consider a relaxed version of the problem, in which parameters are softly constrained to take values close to their average cluster values. To achieve this, we choose the regularizer function $R$ to be  a clustering penalty on the parameters, specifically the $k$-means loss $J(\boldsymbol W, \boldsymbol{\mu})$, defined to be the sum of the distance between each parameter and its corresponding cluster center,
\begin{align}
R_{\boldsymbol{\mu}}(\boldsymbol W) \triangleq J(\boldsymbol W, \boldsymbol{\mu}) \triangleq \frac{1}{2} \sum_n \min_k \|w_n - \mu_k \|_2^2= \frac{1}{2} \sum_k \sum_{i \in C_k} \|w_i - \mu_k\|_2^2  \label{eq:distort}
\end{align}
where $C_k=\{i|k=\arg\min_j \|w_i-\mu_j\|_2^2\}$ contains indices of parameters in cluster $k$, and $\boldsymbol{\mu} \in \mathbb{R}^K$ is the vector of cluster centers.  Note that $J$ defines a shifted $\ell_2$ norm without the restriction $\boldsymbol \mu$=0. From a Bayesian view, given a fixed clustering, $J(\cdot, \boldsymbol{\mu})$ represents a prior over the parameters that consists of $K$ independent Gaussian components with different means and shared variances. 

While $k$-means has been used for parameter quantization after training \citep{han2016,gong2014}, we propose to incorporate it directly into the objective as a prior. The hope is that this prior will guide the training towards a \textit{good} parameter tying from which hard-tying (i.e., enforcing the parameter tying constraints) will incur a relatively small loss. Indeed, one of the main observations of this paper is that the $k$-means prior \eqref{eq:distort} proves to be highly effective for inducing quantization. 

The $k$-means prior has fewer parameters/hyperparameters to learn/tune compared to a GMM prior; in addition, it is more natural if we believe that the data is actually generated from a model with finitely many distinct parameters: we expect both priors to perform comparably when the distinct parameters are far apart from each other, but as the clusters move closer together, the GMM prior leads to clusters with significant overlap. In the worst case, the GMM prior converges to a mixture such that each parameter has almost exactly the same probability of being generated from each mixture component.  This yields poor practical performance.  In contrast, $J$ forces each parameter to commit to a single cluster, which can result in a lower loss in accuracy when hard-tying. In addition, the maximum likelihood objective for the GMM prior can encounter numerical issues if any of the variances tends to zero, which can happen as components are incentivized to reduce variances by eventually collapsing onto the network parameters. This problem can be alleviated by setting individual learning rates for the GMM and model parameters, annealing the GMM objective \citep{nowlan1992}, or imposing hyperpriors on the GMM parameters to effectively lower-bound the variances \citep{ullrich2017}; still, significant computation and tuning may be required for good solutions.

\section{(Sparse) Automatic Parameter Tying}

Following the approach of \citet{han2016}, if we store the original parameters of a model using $b$-bit floats (typically 16 or 32) and quantize them so that they only take $K$ distinct values, then we only need to store the cluster means, $\boldsymbol{\mu}$, in full precision and the quantized parameters by their index, corresponding roughly to a compression rate of \begin{align}
r = \frac{Nb}{N \log_2{K} + Kb}. \label{eq:comp}
\end{align}
For a parameter-heavy model such that $N \gg K $, the denominator in \eqref{eq:comp} is dominated by $N \log_2{K}$, so most of the savings from quantization comes from storing parameter indices with $\log_2{K}$ instead of $b$ bits.
However, quantization alone has its limitations: for example, if $b=32$, a high compression rate as computed in  \eqref{eq:comp}, e.g., a rate of over $100$, would practically require $K=1$ (entire network with a single parameter value), which is infeasible without high accuracy loss. 

To reduce the number of parameters that need to be explicitly stored, we consider another common strategy for compression, network pruning, which results in sparse parameterizations that can be efficiently stored and transmitted using sparse encoding schemes. 
Here, we use the scheme proposed by \citet{han2016} and detailed by \citet{ullrich2017}, in which parameters are first stored in regular CSC or CSR format and then further compressed by Huffman coding. 
Although network pruning is generally orthogonal to quantization, we can achieve both by encouraging a large cluster near zero (referred to as the \textit{zero cluster}): parameters in the zero cluster which are effectively zero can be dropped from the model, and neurons that have only zero weights can also be dropped. To this end, we add an additional sparsity-inducing penalty $E_S(\boldsymbol W)$ to the learning objective 
resulting in the joint learning objective,
\begin{equation}
\min_{\boldsymbol{W},  \boldsymbol{\mu}} E_D(\boldsymbol{W}) + \lambda_1 J(\boldsymbol{W}, \boldsymbol{\mu}) + \lambda_2 E_S(\boldsymbol W),\label{eq:obj}
\end{equation}
The case in which $\lambda_2=0$, corresponding to no sparsity inducing prior, will be simply referred to as APT (Automatic Parameter Tying) or \textit{plain} APT; the other case as \textit{sparse} APT.  In this work, we consider the lasso penalty $E_S(\boldsymbol W) \triangleq {||\boldsymbol W||}_1$, and find experimentally that this additional penalty increases model sparsity without significant loss in accuracy, for large enough $K$. 


We propose a two-stage approach to minimize \eqref{eq:obj}.  In stage one, soft-tying, the objective is minimized using standard gradient/coordinate descent.  In stage two, hard-tying, the soft clustering penalty is replaced with a hard constraint that forces all parameters in each cluster to be equal (parameters in the zero cluster must be zero for sparse APT); the data loss is then minimized using projected gradient descent.  Unfortunately, \eqref{eq:obj} is not a convex optimization problem, even if $E_D$ is convex, as the $K$-means objective $J$ is not convex, so our methods will only converge to local optima in general.

\subsection{Soft-Tying (Coordinate Descent)}
We propose to optimize the (sparse) APT objective $\mathcal{L}$ \eqref{eq:obj} with a simple block coordinate descent algorithm that alternately optimizes with respect to $\boldsymbol \mu$ and $\boldsymbol W$.

Given $\boldsymbol W$, optimization w.r.t to $\boldsymbol \mu$  is solved precisely by the $k$-means algorithm. We consider $\mathbf{C}$ a separate variable (as in standard EM-style $k$-means), and only optimize w.r.t it infrequently for efficiency, instead of eagerly according to its definition; i.e., between every coordinate update to parameters $\boldsymbol W$, we only update cluster centers $\boldsymbol \mu$ (but not $\mathbf{C}$), and only run the full $k$-means procedure to update both $\mathbf{C}$ and $\boldsymbol \mu$ once every 1000 or so parameter updates. As we show in experiments, the frequency of $k$-means updates does not significantly impact the results.
Given $\boldsymbol \mu$ (and $\mathbf{C}$), optimizing w.r.t. $\boldsymbol W$ involves ordinary gradient descent on $\mathcal{L}$ using backpropagation, with weight decay from $J$ \eqref{eq:distort} driving parameters towards their respective cluster centers (as well as $\ell_1$ penalty in sparse APT).

\subsection{Hard-Tying (Projected Descent)}
Once the combined objective has been sufficiently optimized, we replace soft-tying with hard-tying, during which the learned clustering assignment $\mathbf{C}$ is fixed, and parameters are updated subject to tying constraints imposed by $\mathbf{C}$.
Prior to hard-tying, the tying constraints are enforced by setting parameters to their assigned cluster centers; for sparse APT, we also identify the zero cluster as the one with the smallest magnitude, and create sparsity by setting it to zero. 
 
In hard-tying, we optimize the data loss $E_D$ via projected gradient descent (the $\ell_1$ loss in soft-tying with sparse APT is dropped in hard-tying): the partial derivatives are first calculated using backpropagation and then all components of the gradient corresponding to parameters in cluster $k$ are set to their average to yield the projected gradient update.
We note that this is distinct from \citet{han2016}, which updates a cluster center by the sum of partial derivatives of parameters in that cluster instead of the average. This difference arises as \citet{han2016} only allows parameter sharing within each layer, while our projected gradient method handles parameter tying across layers.

\subsection{Computational Efficiency}
We note that unlike the GMM penalty \citep{nowlan1992} the $K$-means problem can be solved exactly in polynomial time in the one-dimensional (1-D) case using dynamic programming \citep{wang2011}, though it isn't particularly efficient in practice. In our implementation, we sped up standard $k$-means by specializing it to 1-D: we take advantage of the fact that comparison in 1-D can be done on entire sets of parameters, if we sort them in advance, and operate on partitions of parameter clusters that implicitly define cluster membership. Thus optimizing the cluster assignments reduces to binary searching between neighboring partitions for partition means, in order to redraw cluster boundaries (in $O(K\log N)$ time), and optimizing the partition means given assignments takes $O(N)$ time, but can be greatly reduced by caching the partition statistics. For the $k$-means steps in soft-tying, we did not observe significant difference in the learning outcome between the dynamic programming $k$-means \citep{wang2011} and our fast approximate 1-D $k$-means (fixing the number of iterations to 100), so we employ the latter approach in all of our experiments. 
 
Finally, we note that our method adds little overhead to network training. The memory requirement is $O(N)$, as the cluster assignments $\mathbf{C}$ are stored as an $N$-vector of integers. 
The computation of cluster means after each gradient step takes linear time $O(N)$, which adds little to the cost of standard back-propogation. The additional $k$-means steps in soft-tying also adds at most $O(N)$ time to the entire training procedure, where the constant term is small as they are run only infrequently.

\section{Experiments}

We used Tensorflow \citep{tensorflow} to optimize \eqref{eq:obj} with respect to $\boldsymbol W$.
For learning the clustering through soft-tying, we implemented the 1-D version of $k$-means in C++ for efficiency, although $k$-means is also provided in standard scientific computing libraries.
In fact, soft-tying can be directly done by SGD and auto-differentiation with a neural network library, but naive computation for $J \eqref{eq:distort}$ requires $O(NK)$ time/memory, so is not used (the results are comparable to our 1-D $k$-means with coordinate descent). 
We implement hard-tying by first updating $\boldsymbol W$ with $\nabla_{\boldsymbol W}E_D$ as usual and then projecting $\boldsymbol W$ onto the constraints imposed by the learned cluster assignments, i.e., setting each parameter to its cluster average; for sparse APT we also keep parameters in the zero cluster at zero.

Unless otherwise specified, we initialize the neural network parameters using the method proposed by \citet{glorot2010}, and initialize the cluster centers heuristically by evenly distributing them along the range of initialized parameters. 
As our experiments are concerned with classification problems, we use the standard cross-entropy objective as our data loss.
In experiments with MNIST and CIFAR-10 image datasets, 
we use the original train/test split provided, form a validation set from 10\% of training data, and normalize the data by mean/variances of the training set. 

We present three sets of experiments.  First, we perform APT on MNIST to examine the effect of the $k$-means prior and associated learning dynamics. 
Inspired by recent work on neural network generalization, our second set of experiments on a CNN and its locally-connected version aims to understand the generalization effect of APT and parameter tying in general. Our last set of experiments compares the compression performance of sparse APT and other state-of-the-art methods.

\subsection{Algorithmic Behavior}

\begin{figure}[t]
    \centering
    \begin{minipage}{0.30\textwidth}
        \centering
        \includegraphics[width=0.95\textwidth]{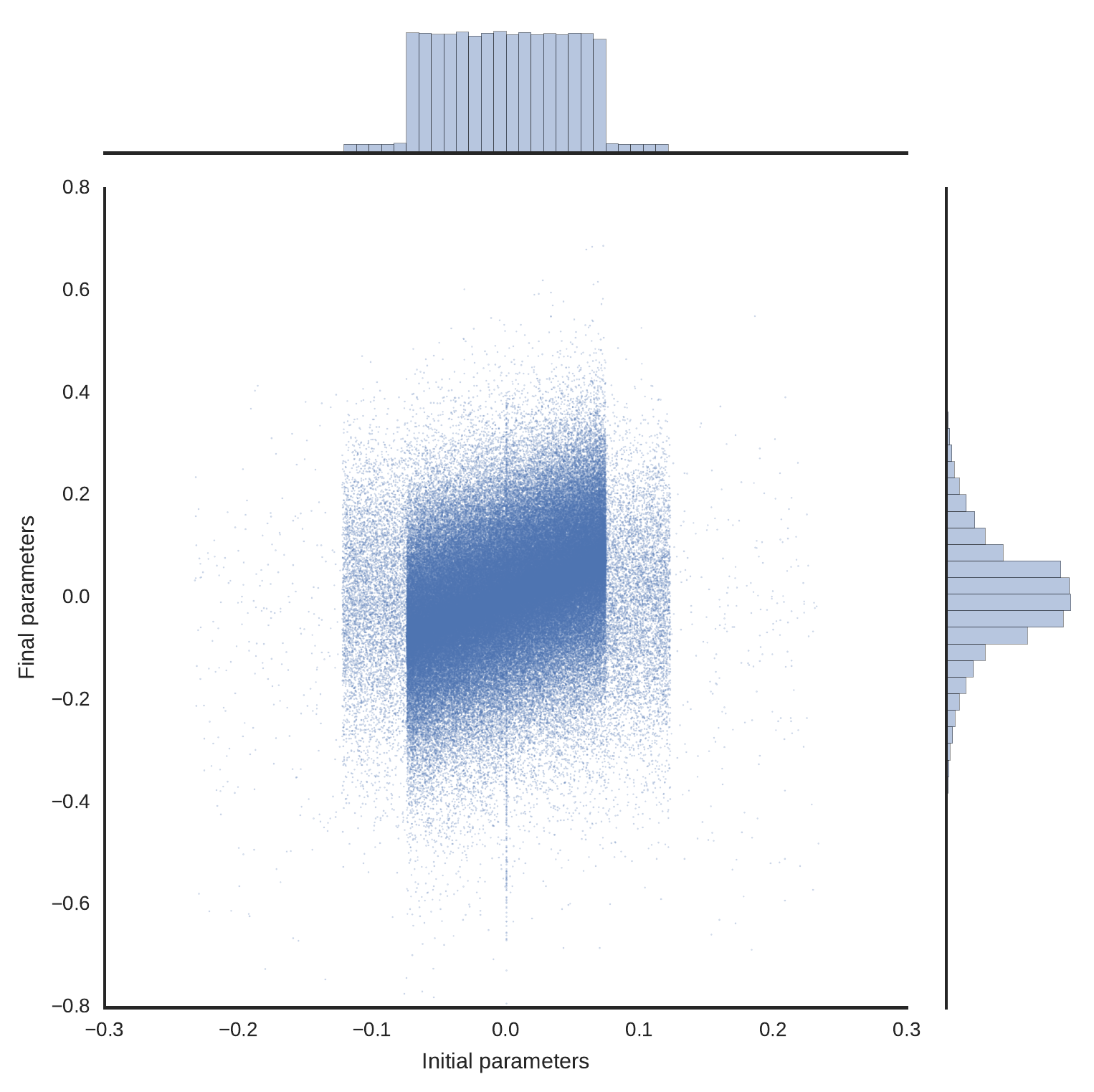} 
    \end{minipage}\hspace{1cm}
    \begin{minipage}{0.30\textwidth}
        \centering
        \includegraphics[width=0.95\textwidth]{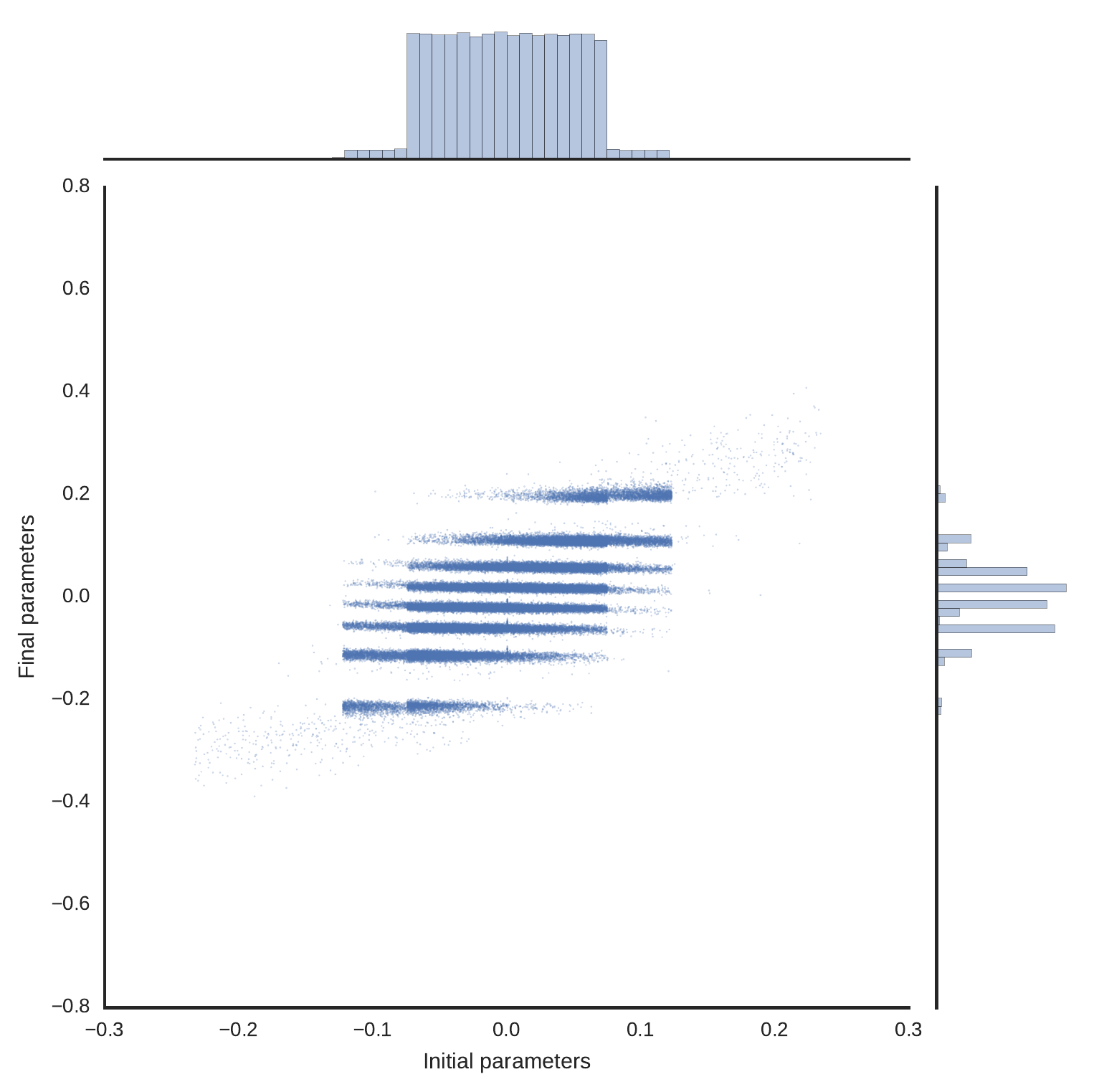} 
    \end{minipage}
    \caption{Joint histograms of parameters before and after training, without (left) and with an additional $k$-means loss (soft-tying APT). The parameters are initialized with scaled uniform distributions proposed in \citep{glorot2010} and $K=8$.}
\label{fig:hist}
\end{figure}
\begin{figure}[t]
\centering
    \subcaptionbox{Cross-entropy with $k$-means loss.
    \label{fig:apt_losses}}			{\includegraphics[width=.40\textwidth]{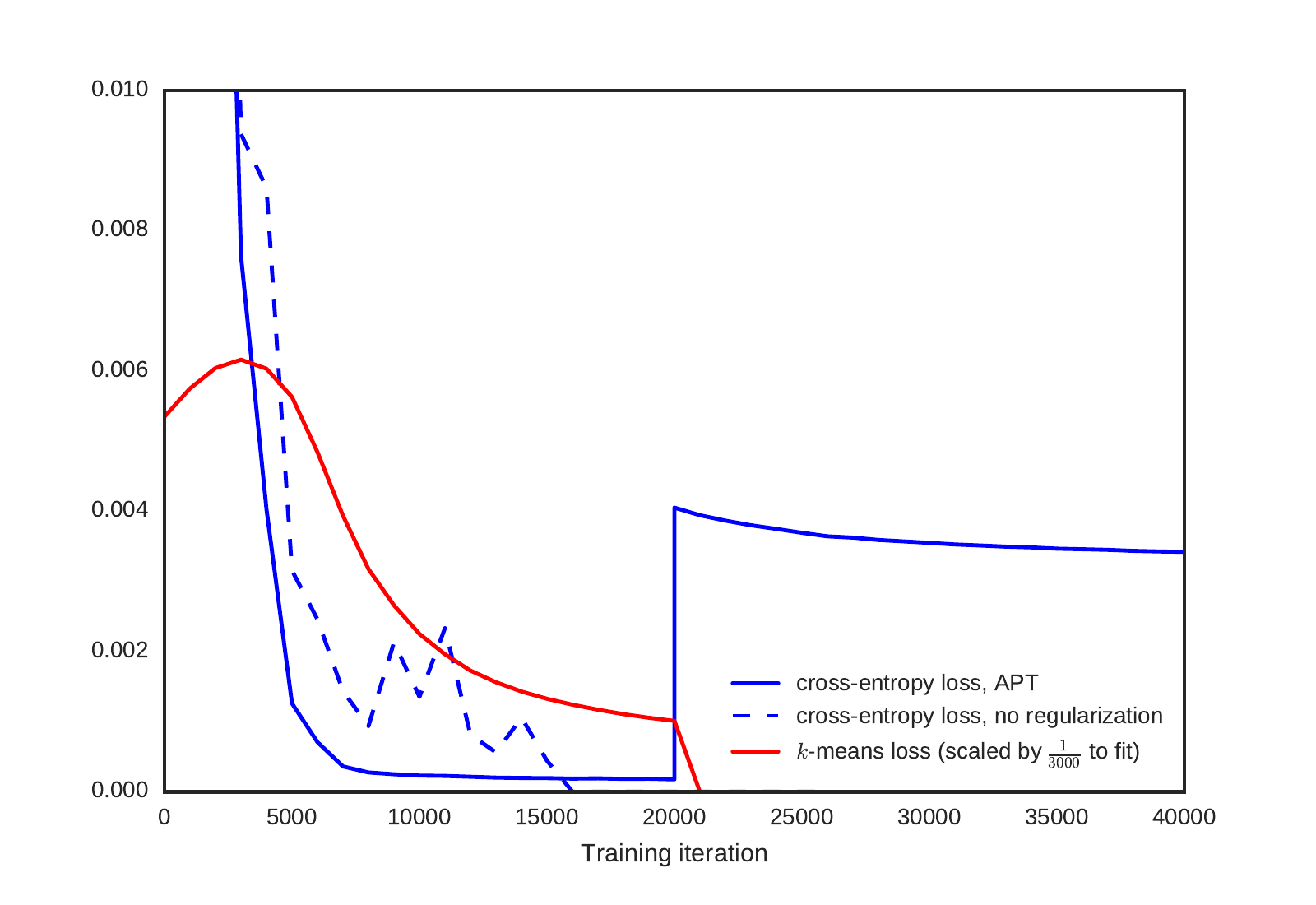}} 
 \subcaptionbox{Error rates.\label{fig:apt_errs}}
 {\includegraphics[width=.40\textwidth]{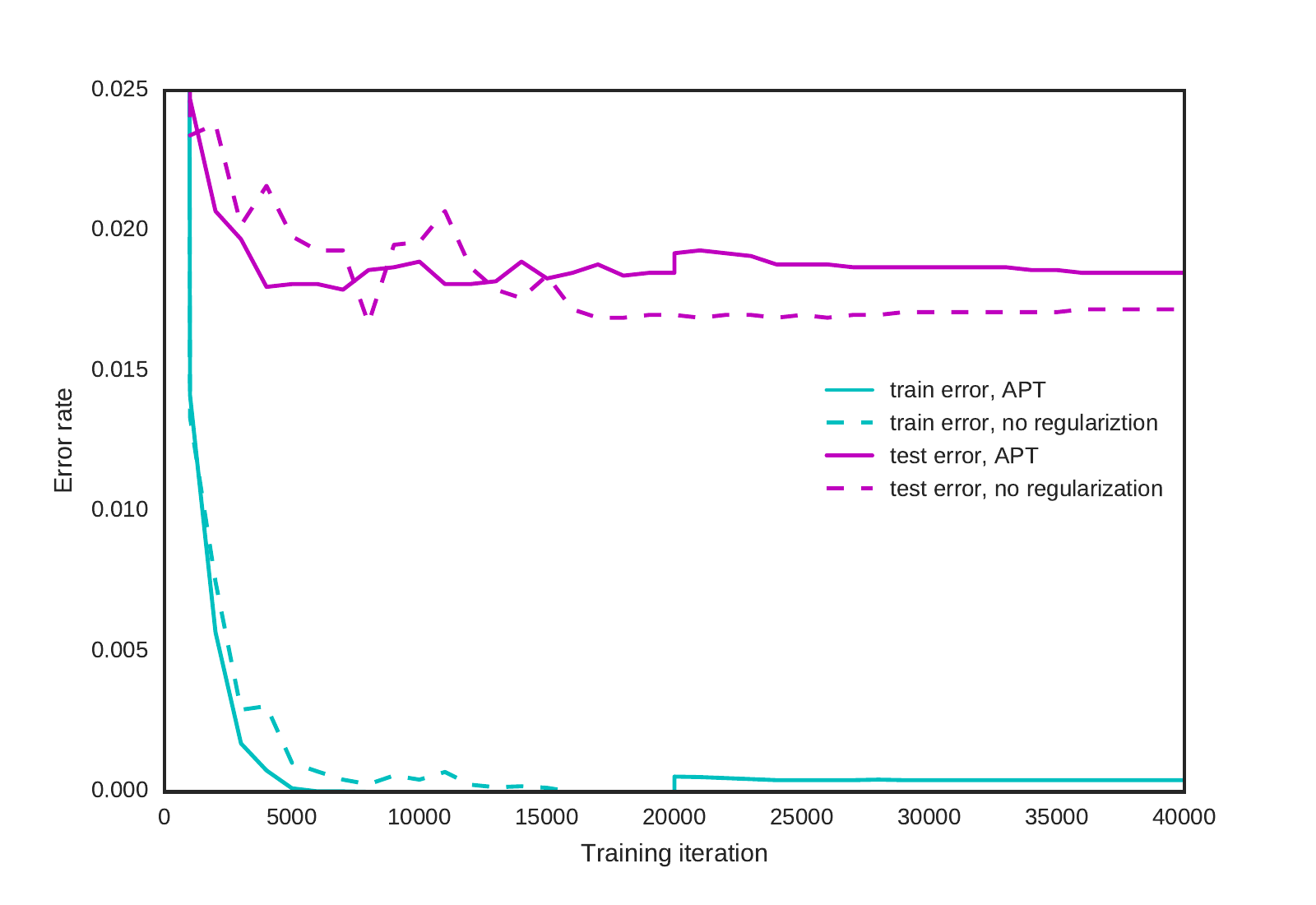}} 
 \caption{Comparison of training with APT (first 20000 iterations soft-tying, last 20000 hard-tying) vs. without regularization on LeNet-300-100, using the same initialization/learning rate.} \label{fig:apt_losses_errs}
\end{figure}

We demonstrate the typical behavior of APT using LeNet-300-100 on MNIST.  We trained with soft-tying for 20000 iterations, and switched to hard-tying for another 20000 iterations. Figure \ref{fig:hist} depicts a typical parameter distribution produced by APT at the end of soft-tying versus training without any regularization, using the same initialization and learning rate. As expected, APT leads to a clear division of the parameters into clusters. Figure \ref{fig:apt_losses_errs} illustrates the loss functions and model performance in the experiment, with and without APT. In this demonstration, $K$=8 appeared sufficient for preserving the solution from soft-tying: switching from soft to hard-tying at iteration 20000 resulted in some small loss, and hard-tying was able to gradually recover from it. 
Generally for a properly chosen $K$, soft-tying does not fundamentally change the convergence speed or final model performance, compared to without APT. 
However, the loss in accuracy from hard-tying can be significant for small $K$, and decreases with increasing $K$.  The hard-tying phase is generally able to recover from some or all of the accuracy loss for large enough $K$.  See Appendix \ref{app:exp} for details.

We also explored in Appendix \ref{app:exp} the effect of coordinate switching frequency on the learning outcome, for which we reran the previous experiments with varying frequency of $k$-means steps. We observed that APT was generally insensitive to $k$-means frequency, except for very small $K$, justifying our heuristic of only running $k$-means infrequently. 
We also observe that random tying is disastrous for small $K$, which simply can't effectively cover the range of parameters and induces significant quantization loss. Although special techniques exist for training networks with $K$=2 or 3, e.g. \citep{cour2016}, our current formulation cannot effectively quantize at this level.  


\subsection{Effect on Generalization}
Recently, \citet{zhang2016} observed that the traditional notion of model complexity associated with parameter norms captures very little of neural networks' generalization capability: traditional regularization methods, like 
$\ell_2$ (weight decay), do not introduce fundamental phase change in the generalization capability of deep networks, and bigger gains can be achieved by simply changing the model architecture rather than tuning regularization.
The paper left open questions of how to \textit{correctly} describe neural network's model complexity, in a way that reflects the model's generalization.  In this section, we explore a different notion of model complexity characterized by the number of free parameters in parameter-tied networks, where the tying is discovered through optimization. For demonstration, we present experiments on MNIST where no significant regularization effect of parameter tying was observed, similar to traditional regularization methods; this suggests that enforcing parameter-tying constraints does not constitute a major change in network architecture. Our more extensive experiments (not presented here) point to the same conclusions.

Two of the main architectural features of a CNN are \textit{local connectivity} and \textit{parameter tying}; local receptive fields allow units to extract elementary visual features of images, and tying the weights of all units of a feature map allows detection of a useful feature across an entire image \citep{lecun1998}. 
In an attempt to better understand the regularization/generalization impact of parameter tying and local connectivity in CNN, we explored alternative parameter tying and regularization methods on a locally connected network (LCN) that is identical to CNN but without parameter tying constraints, and similarly on an equivalent feedforward network (MLP) capable of simulating the LCN/CNN. To ensure that the prior assumptions of CNN are met (which may not always be; e.g., LCNs are used for face recognition \citep{taigman2014}), we use the MNIST dataset as in the original CNN paper \citep{lecun1998}.
We chose the popular LeNet-5-Caffe architecture as the reference CNN, and trained the corresponding LCN and MLP with either  no regularization (``no reg''), $\ell_1$ regularization, $\ell_2$ regularization, or APT. With APT, we only tie the locally connected layers of the LCN (and corresponding layers of the MLP) in order to compare with CNN.
All methods were trained to convergence within a max budget of 20000 iterations; for APT, we perform hard-tying for another 10000 iterations after the initial 20000 iterations of soft-tying. We set the parameters of the methods by grid search on the validation set (except that $K$ was set on log scale for APT simply for illustration), and report the corresponding test error, averaged over 3 random runs (the standard deviations were roughly the same for all methods and hence not shown).

\begin{figure}[t]
\centering
\begin{minipage}{.49\textwidth}
  \centering
  \includegraphics[width=.7\linewidth]{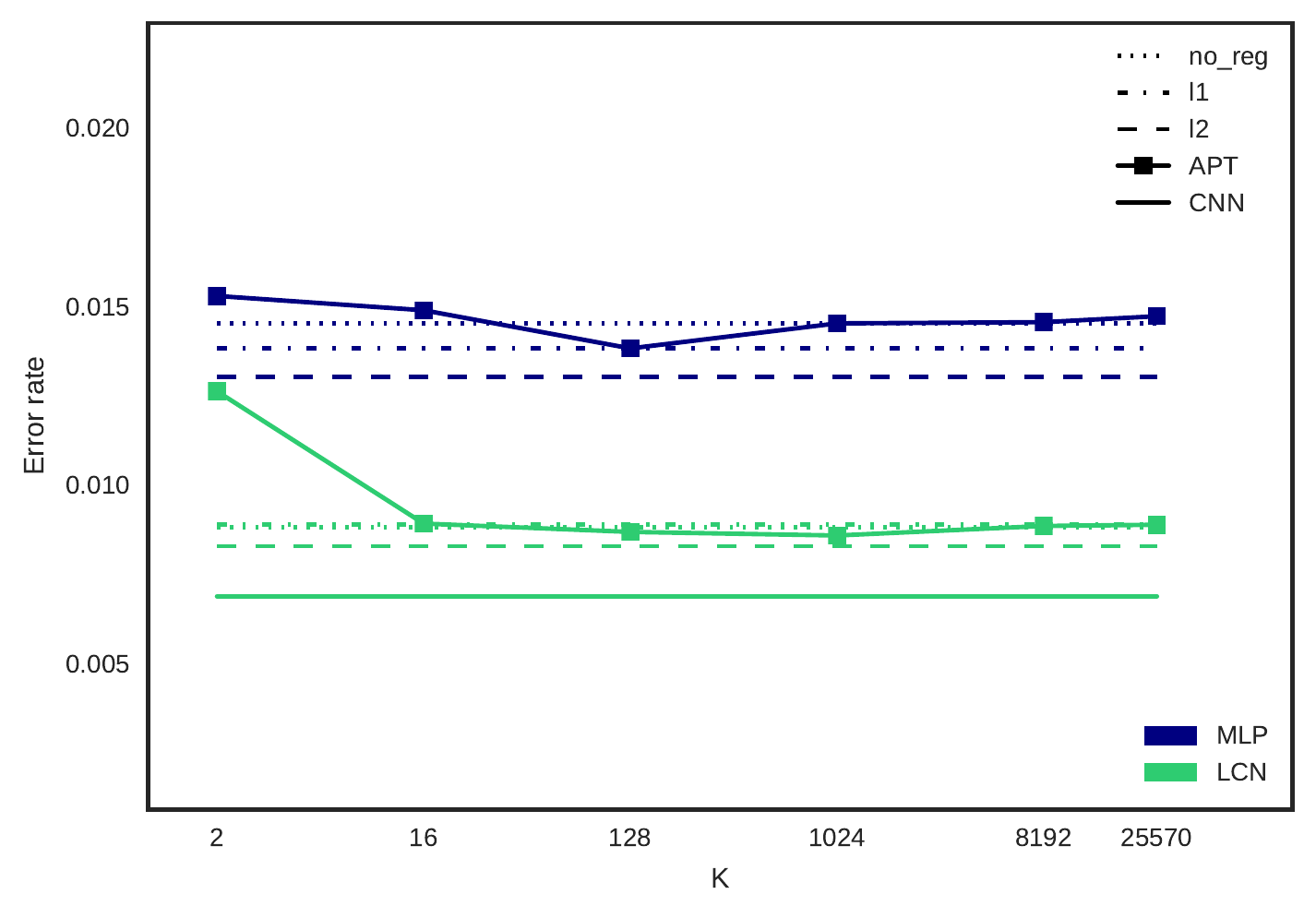}
\end{minipage}%
\begin{minipage}{.49\textwidth}
  \centering
  \includegraphics[width=.75\linewidth]{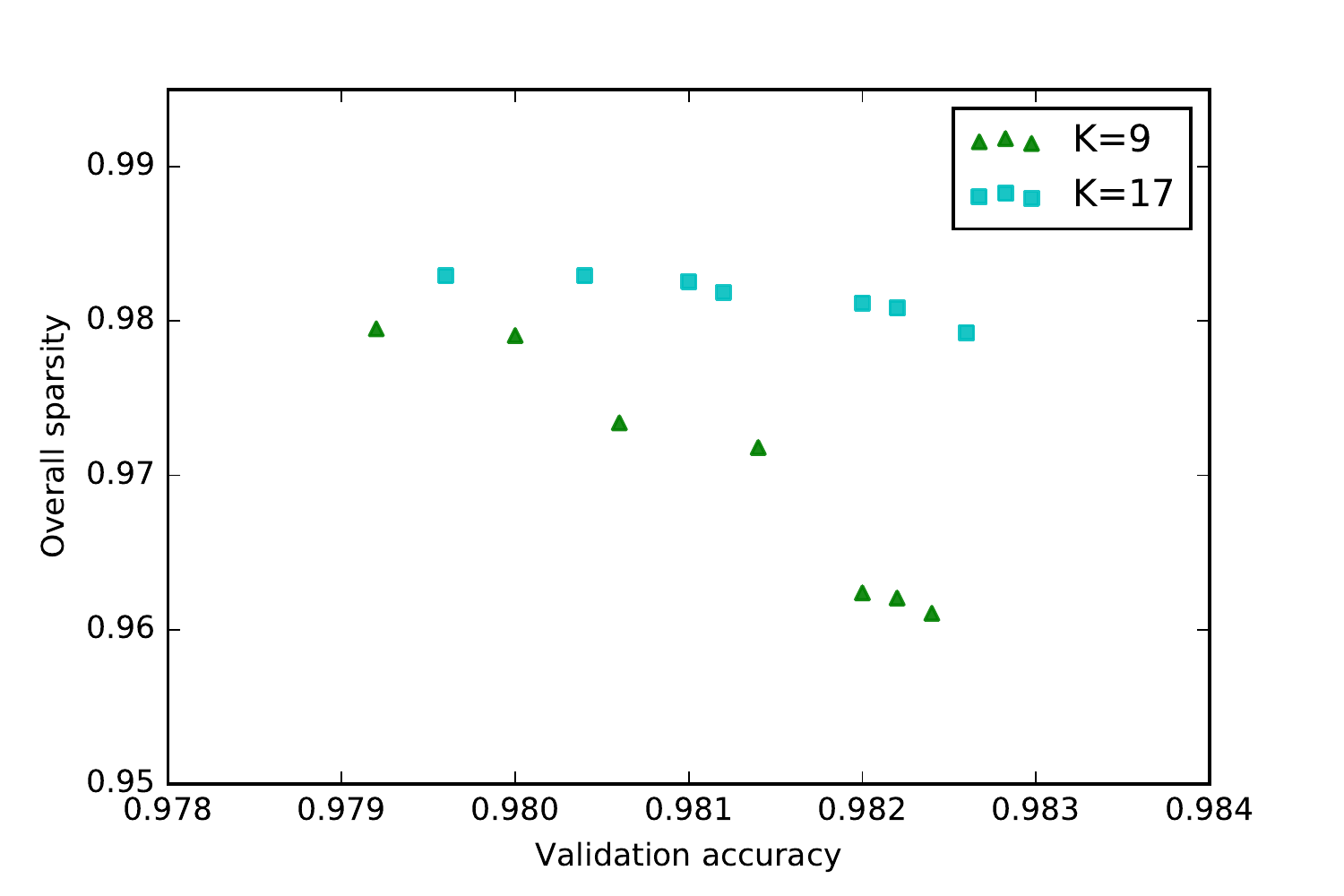}
\end{minipage}
\par
\medskip
\noindent
\begin{minipage}[t]{.48\textwidth}
  \centering
  \captionof{figure}{Test errors of equivalent LCN and MLP trained with various regularization methods (with increasing $K$ for APT) compared to CNN. 
  }
  \label{fig:lcn_err}
\end{minipage}%
 \hfill
 \begin{minipage}[t]{.49\textwidth}
  \centering
  \captionof{figure}{Sparsity versus accuracy trade-off for LeNet-300-100, shown as the Pareto frontier of typical hyper-parameter search results.}
  \label{fig:pareto}
\end{minipage}
\end{figure}



As can be seen in Figure \ref{fig:lcn_err}, parameter-tying with APT resulted in no significant loss in accuracy for $K>2$; additionally there was no noticeable performance difference for values of $K$ between 4 and 25570, the number of distinct parameters in the CNN convolution layers.
Parameter tying (either through APT or convolution) appears to belong with the other explicit regularization methods, in that they all achieved essentially the same performance (not significantly better than without regularization).
Note that \citet{zhang2016} also place data augmentation and dropout in this category.
Switching the parameter tying scheme from APT (or none at all) to CNN reduced the error rate by about 0.001, or 0.1\%, which is insignificant compared to \textit{changing the network architecture from fully-connected to locally-connected}, which reduced error by 0.5\% $\sim$ 0.7\%. Despite similar performance of APT on the LCN compared to the CNN, we found that APT did not recover the ``ground-truth'' parameter tying of CNN, which constrains all the local filters associated with a feature map to be identical. A visualization of LCN filter can be found in Figure \ref{fig:lcn_weights_vis} in Appendix \ref{app:weight_vis}.


\subsection{Sparsity and Compression Results}
\begin{table*}[t]

\caption{Comparison of sparse APT with other compression and/or sparsity-inducing methods.}
\label{tab:cr_results}
\centering 
\begin{footnotesize}
\begin{tabular}{clcccc}
\specialrule{.1em}{.05em}{.05em} 
Network  & Method & Error \% & $\frac{|w\neq 0|}{|w|}\%$ & Max. Compression Rate\\
\hline
LeNet-300-100         & DC & 1.6 & 8.0 & 40\\
& SWS & 1.9 & 4.3 & 64  \\
& Sparse VD & 1.8 & 2.2 & 113 \\
& BC-GNJ & 1.8 & 10.8 & 58 \\
& BC-GNS & 2.0 & 10.6 & 59 \\
& Sparse APT & 1.9 & 2.1 & 127 \\
& Sparse APT (DC) & 1.6 & 3.6 & 77 \\

\hline
LeNet-5-Caffe         & DC & 0.7 &  8.0 & 39 \\
& SWS & 1.0 & 0.5 & 162 \\
& Sparse VD & 1.0 &  0.7 & 365 \\
& BC-GNJ & 1.0 & 0.9 & 572 \\
& BC-GNS & 1.0 & 0.6 & 771 \\
& Sparse APT & 1.0 & 0.5 & 346\\
& Sparse APT (DC) & 0.7 & 6.9 & 45  \\ 
\hline
VGG-16               & BC-GNJ & 8.6 & 6.7 & 95 \\
& BC-GNS & 9.2 & 5.5 & 116 \\
& Sparse APT & 8.3 & 4.6 & 93 \\
 \specialrule{.1em}{.05em}{.05em} 
\end{tabular}
\end{footnotesize}

\end{table*}
We compare sparse APT against other neural network compression or pruning methods, including Deep Compression (DC) \citep{han2016}, Soft Weight Sharing (SWS) \citep{ullrich2017}, Bayesian Compression (BC) \citep{louizos2017}, and 
Sparse Variational Dropout (Sparse VD) \citep{molchanov2017} using LeNet-300-100 and LeNet-5-Caffe on MNIST, and VGG-16 on CIFAR-10. 
We perform sparse APT by first soft-tying for a fixed budget of iterations and then hard-tying for another budget of maximum iterations. In our experiments, we found that in order to achieve $\leq 1 \%$ accuracy loss, $K$ in [10, 20] was sufficient for networks with several million parameters or less and $K$ in [30, 40] sufficient for 10 to 20 million parameters. We tuned $\lambda_1$ and $\lambda_2$ in $[10^{-6}, 10^{-3}]$ with grid search on log scale and manual tuning. In general we found the $\ell_1$ penalty to have little impact on $k$-means loss \eqref{eq:distort} or cluster convergence, so we could tune $\lambda_2$ independently of a reasonable $\lambda_1$ to control the sparsity level.

For compressing LeNets, we used the Adadelta \citep{zeiler2012} step size rule, no data augmentation or other regularization, and soft/hard-tying budgets of 60000/10000 iterations respectively. Unlike in methods such as SWS and BC, we found no loss of accuracy for similar sparsity levels when training from random initialization compared to from a pre-trained network, using largely the same number of iterations. 
For VGG-16, we used the same amount of data augmentation, dropout, and batch normalization as in \citep{louizos2017}. 
The training was done by SGD with 0.9 momentum in which the initial learning rate, 0.05, decays by half once the validation accuracy does not improve for 10 consecutive iterations.
We observed that training VGG-16 from scratch could not achieve the same accuracy as from a pre-trained network (about 2\% higher error for similar sparsity).
We used soft/hard-tying budgets of 80000/20000 iterations, starting with a pre-trained model with 7.3\% error.

The results are presented in Table \ref{tab:cr_results}. We report the error of the networks on the test set, the fraction of non-zero weights, 
, and the Maximum Compression Rate as in \citep{ullrich2017}. Note that \citet{louizos2017} evaluate the compression criteria separately for each of their variants of BC, instead of with a single trained network, following the sparsity/compression statistics as in \citep{louizos2017}. The Maximum Compression Rates for DC, BC, and Sparse VD were obtained by clustering the final weights into 32 clusters (this achieved the best compression rate \citep{louizos2017}).  SWS used $K$=17 for LeNets, and sparse APT used $K$=17 for LeNets and $K$=31 for VGG-16, corresponding to 16 and 30 distinct non-zero parameter values. When evaluating sparse APT at the same error level as DC on LeNets (1.6\% for LeNet300-100 and 0.7\% for LeNet-5), we found $K$=17 insufficient for achieving such low errors and instead used $K$=33 (the same as in DC); the results are shown under ``Sparse APT (DC)".


Overall, we observe that sparse APT outperforms or performs similarly to all competitors on each data set, with the exception of the BC methods in terms of Max Compression Rate on LeNet-5 and VGG-16; this occurs even though sparse APT manages to find a sparser solution than both BC variants. The explanation for this is that the Maximum Compression score uses 
Huffman coding to compress the cluster indices of quantized parameters in CSR format. As Huffman coding performs best with non-uniform distributions, the primary difference between the sparse APT and the BC solutions is that the BC solutions do not return many equal sized clusters.  While our main goal was to achieve sparsity with a small number of parameters, if a high Maximum Compression Rate is desired, the variances of the independent Gaussian prior could be tuned to induce a significantly more non-uniform distribution which may yield higher compression rates. 

More generally, APT can be used to trade-off between accuracy and sparsity depending on the application, by using a validation set. 
Figure \ref{fig:pareto} illustrates part of the sparsity/accuracy trade-off curve for two different values of $K$.  When $K=9$, sparsity can be increased at a significant loss to accuracy, while at $K=17$, additional sparsity can be gained with only moderate accuracy loss.  In practice, selecting the smallest value of $K$ that exhibits this property is likely to yield good accuracy and compression. In fact, the existence of such a $K$ provides further evidence that, for a fixed structure, sparsity and quantization has little impact on generalization performance.

\section{Conclusions}
We proposed a simple, intuitive, and effective neural network compression algorithm based on quantization and sparsity inducing priors that is competitive with state-of-the-art methods which are often much more complicated and/or expensive.  Our approach adds little overhead to standard network training 
and scales well to larger networks, without significant tuning.  In addition, we offered new empirical evidence based on image data that network architecture and connectivity patterns provide stronger regularization effect than parameter tying or norm restrictions.  

For future work, other forms of clustering priors may be explored for quantization while keeping the optimization efficient. For instance, the $\ell_2$ distance in $k$-means prior \eqref{eq:distort} may be replaced with other metrics (e.g., the case of $\ell_1$ distance yields a clustering problem that can be solved by the $k$-medians algorithm).
Similarly, other sparsity inducing priors than $\ell_1$ may be explored.
More efforts would be required to elucidate the relationship between parameter tying and neural network learning and generalization. It would also be interesting to automatically choose $K$ without a validation set (especially if $K$ is to be tuned layer-wise), possibly by nonparametric Bayesian methods such as DP-means \citep{kulis2012} that jointly learn the right clustering and the number of clusters.



\newpage
\bibliography{paper}
\bibliographystyle{icml2018}

\newpage
\appendix
\section{Additional Experimental Results}
\subsection{APT Experiments}
\label{app:exp}
\begin{figure}[h]
    \subcaptionbox{Cluster centers throughout APT training, shaded by 1 standard deviation of points in the cluster. \label{fig:centers_evolution}}{\includegraphics[width=.42\textwidth]{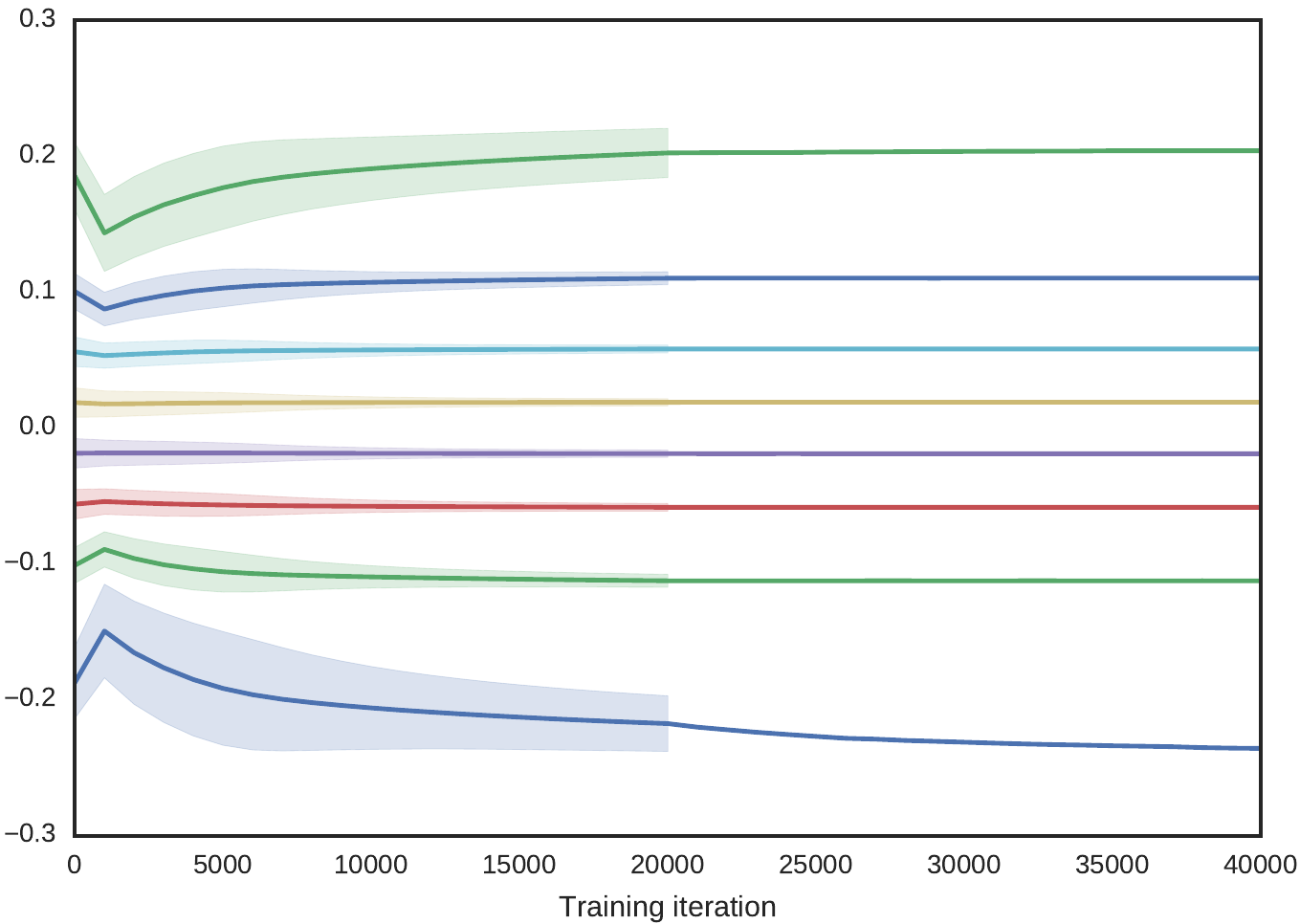}}\hfill
     \subcaptionbox{Change in cluster assignments throughout APT training, as the ratio of parameters that changed assignments in the previous iteration. \label{fig:assignments_evolution}}
        {\includegraphics[width=.5\textwidth]{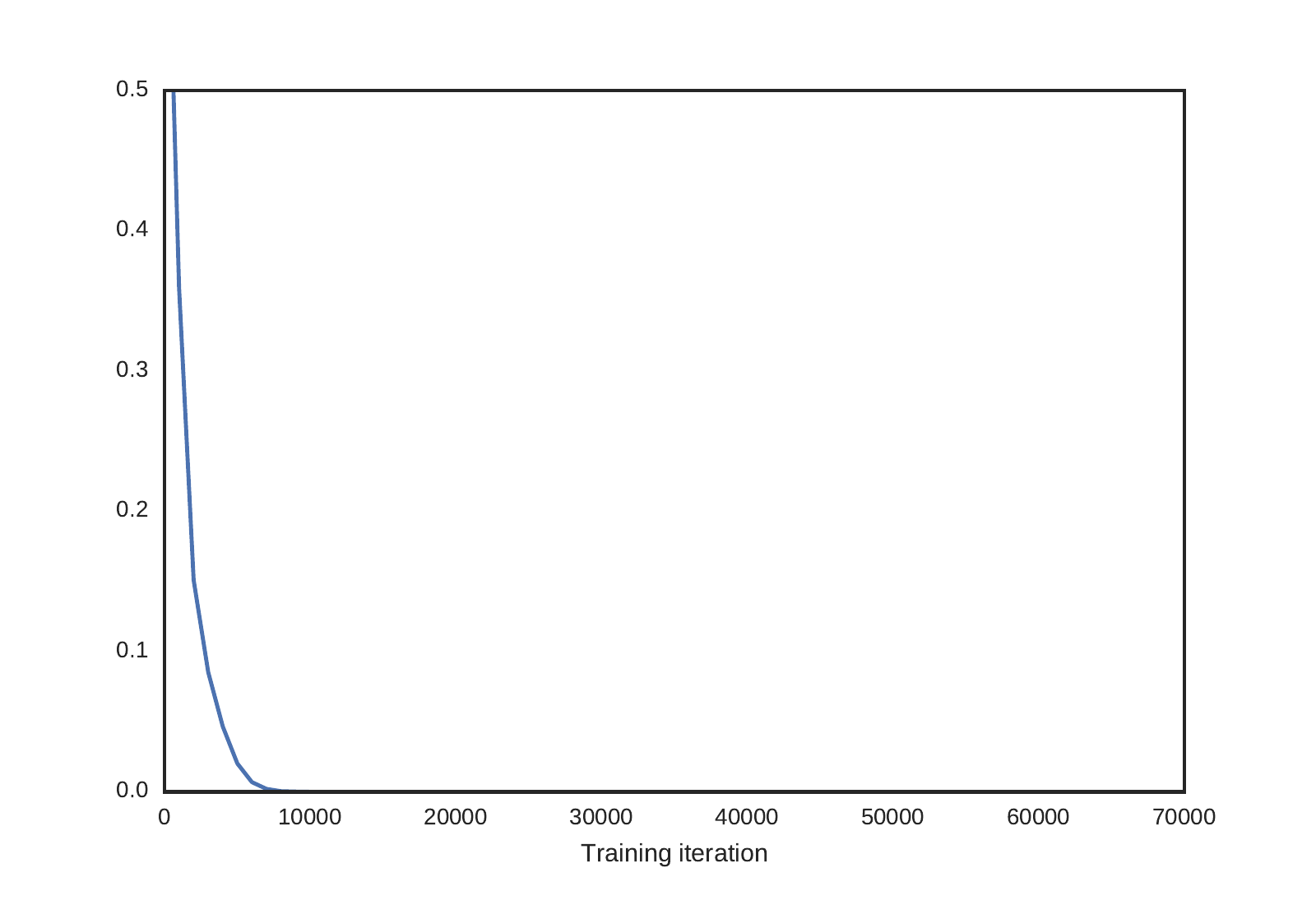}} 
 \caption{Evolution of the clusters in the first APT experiment with LeNet-300-100.} \label{fig:cluster_evolution} 
\end{figure}

Below we illustrate the evolution of cluster centers $\boldsymbol \mu$ and change in cluster assignments $\mathbf{C}$ in the first experiment with LeNet-300-100. Note that the clusters in figure \ref{fig:centers_evolution} tend to oppose  each other, unlike in the case of GMM where they tend to merge; this is a property of $k$-means loss $J$ and independent Gaussian priors. The clusters centers also developed more extreme values during hard-tying.

\begin{figure}[bh]
    \subcaptionbox{Average error rates (along with 1 standard deviation error bars) at the end of soft-tying, and hard-tying, for LeNet-300-100. \label{fig:acc_vs_k_st_ht_errorbars}}{\includegraphics[width=.48\textwidth]{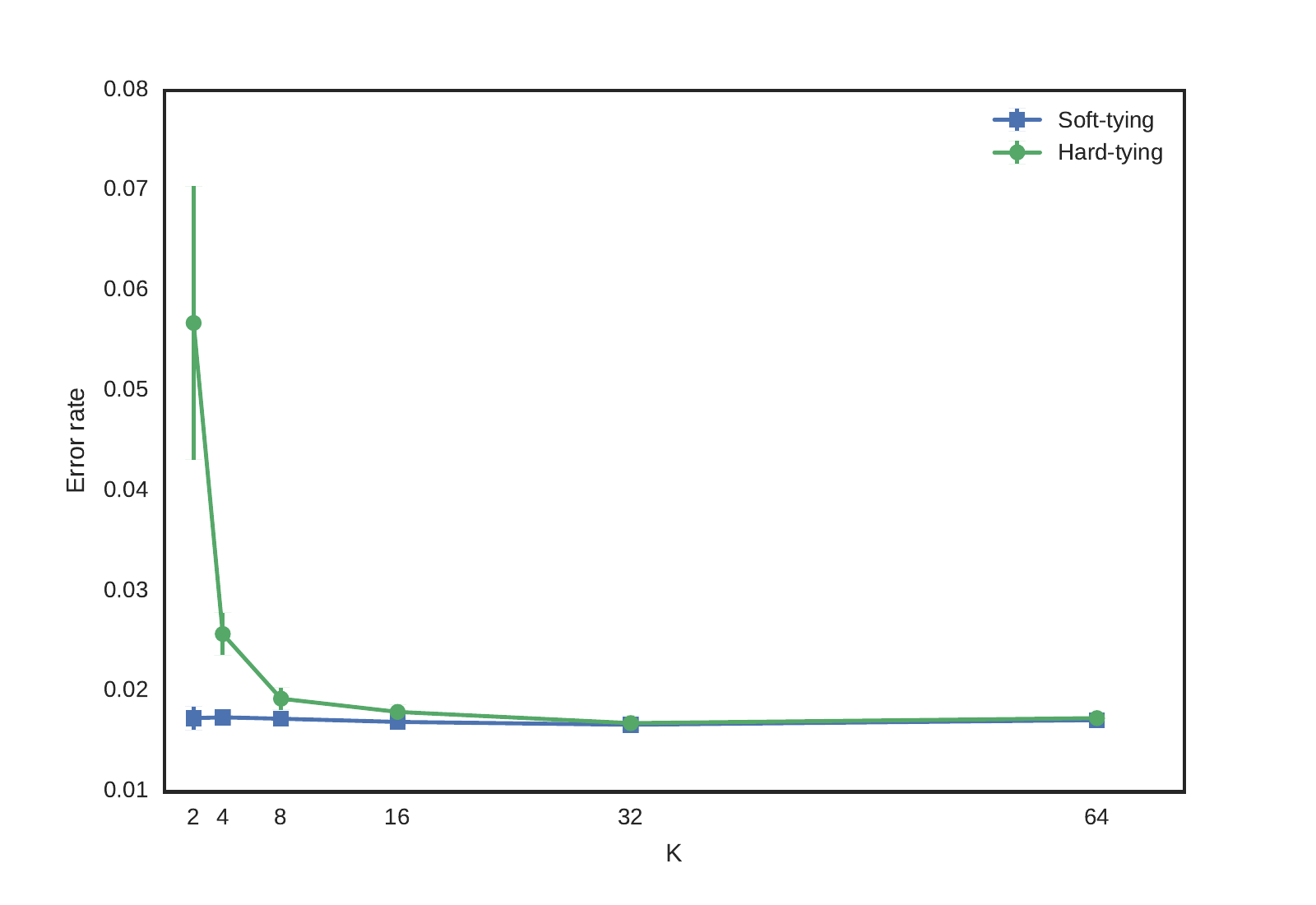}}\hfill
     \subcaptionbox{End of training error rates for various $t$ (number of iterations between $k$-means), for various $K$\label{fig:acc_vs_q_ht}}
        {\includegraphics[width=.48\textwidth]{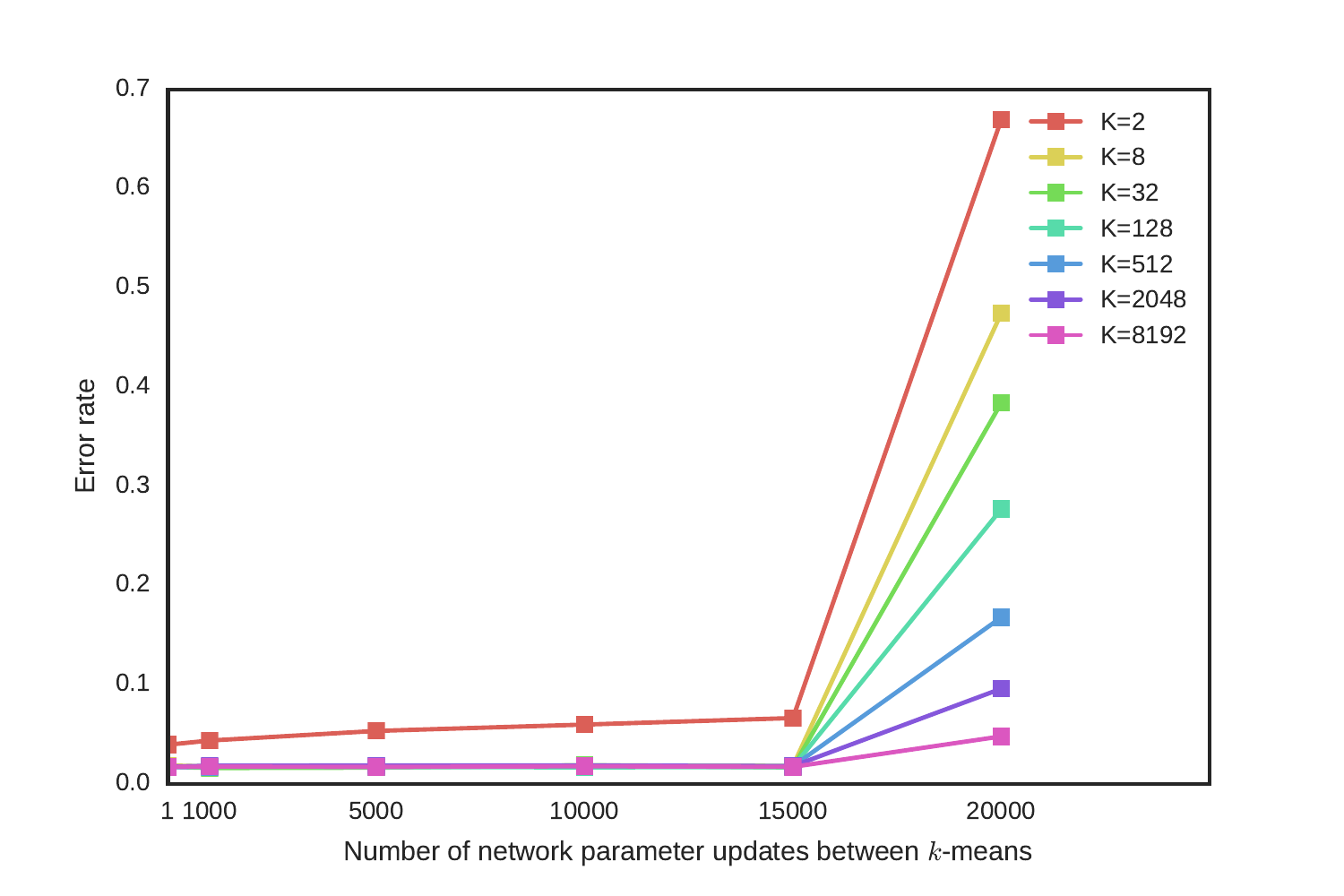}} 
 \caption{Effect of varying $K$ and $t$ on learning outcome with APT.} \label{fig:vary_k_t} 
\end{figure}

We also examined the effect of $K$ with a series of experiments using LeNet-300-100, in which we learned parameter-tied networks with $K = 2, 4, 8, 16,$ and $32$. We ran soft-tying till convergence for a budget of 30000 iterations, followed by another 20000 iterations of hard-tying. We tuned $\lambda_1$ in the range of \{1e-7, 1e-6, ..., 1e-1\}, and selected the best model for each $K$ based on validation performance. We did not observe overfitting with either soft-tying or hard-tying, so for simplicity we considered the model performance at the end of their budgeted runs in each phase.
Figure \ref{fig:acc_vs_k_st_ht_errorbars} displays the best error rates at the end of soft-tying and hard-tying, averaged across 5 random seeds. As can be seen, $K$ did not significantly affect the solution quality from soft-tying; however the accuracy loss involved in switching to hard-tying becomes significant for small enough $K$s, and decreases to zero for $K=32$. 

In another set of APT experiments with similar setup, we examined the impact of $k$-means frequency on model performance for various $K$, in which we vary the number of gradient iterations $t$ between $k$-means runs, with $t \in \{1, 1000, 5000, 10000, 15000, 20000\}$. Soft/hard-tying were set at 20000/20000 iterations. Here we consider the best end of training (soft-tying followed by hard-tying) error rates after hyperparameter search. As can be seen in \ref{fig:acc_vs_q_ht}, $t$ does not appear to be a sensitive hyperparameter, but model performance does degrade with large $t$, particularly for smaller $K$. 
Note that the extreme case of $t=20000$ corresponds not running $k$-means, and hence not updating parameter assignments at all, therefore randomly tying the parameters based on their random initial assignments; this generally prevents effective learning except when $K$ is large.

\subsection{Weight Visualizations}
\label{app:weight_vis}

\subsubsection{APT on LCN}
\begin{figure}[t]
\centering
\begin{minipage}{.49\textwidth}
  \centering
  \includegraphics[width=0.6\linewidth]{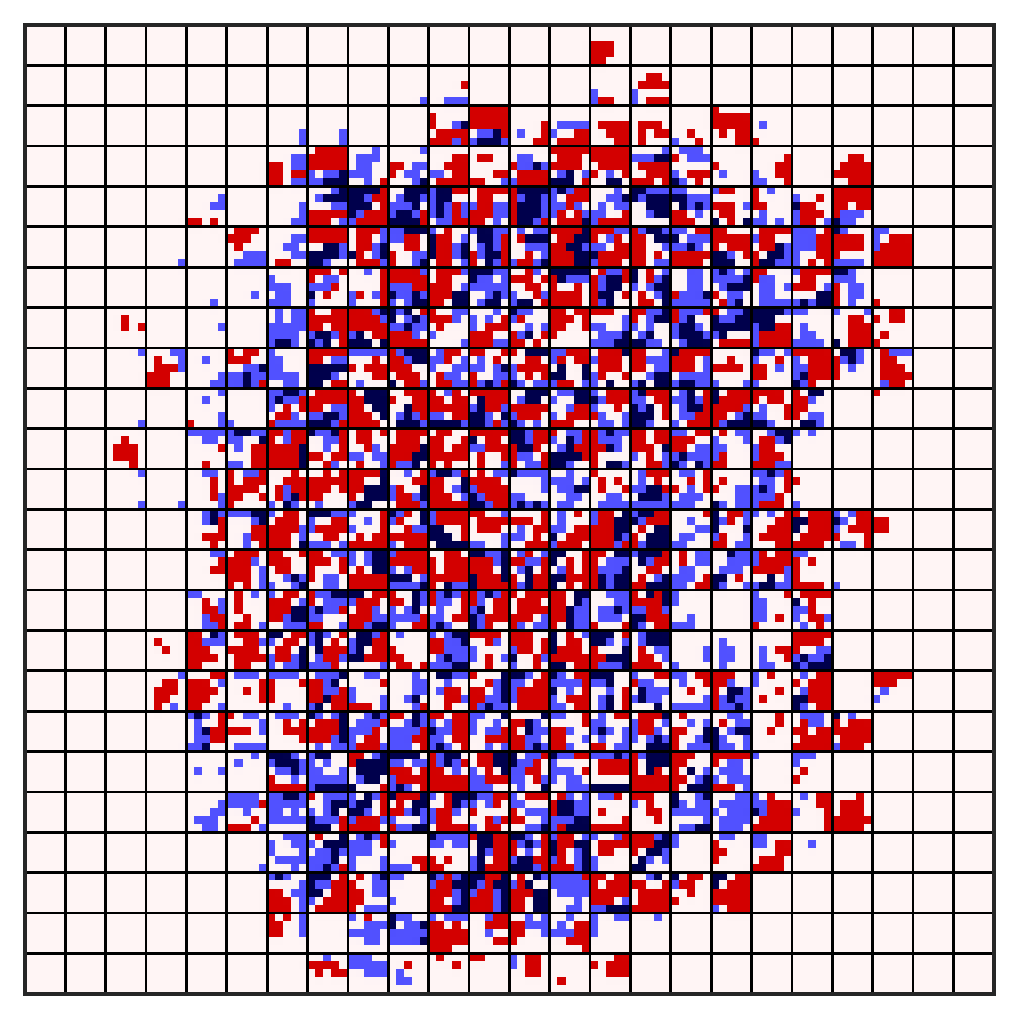}
\end{minipage}%
\begin{minipage}{.49\textwidth}
  \centering
  \includegraphics[width=0.6\linewidth]{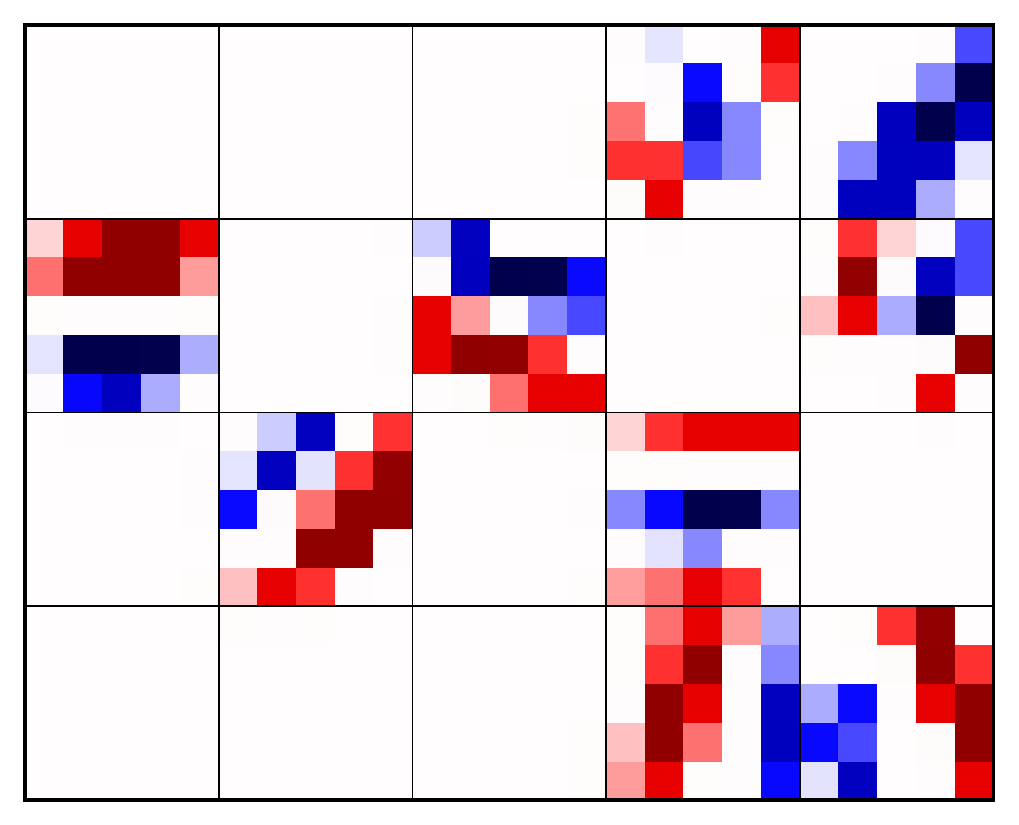}
\end{minipage}
\par
\medskip
\noindent
\begin{minipage}[t]{.48\textwidth}
  \centering
  \captionof{figure}{Visualization of a filter in the first locally connected layer of LCN learned by APT, using $K$=4 and LeNet-5-Caffe as reference CNN, achieving test error rate of 0.0081. Weights in each local filter are arranged into 5 $\times$ 5 grids. In the reference CNN, all of the local filters would share weights and be identical.
  }
  \label{fig:lcn_weights_vis}
\end{minipage}%
 \hfill
\begin{minipage}[t]{.46\textwidth}
  \centering
  \captionof{figure}{Visualization of the first conv layer in LeNet-5, which achieved 1\% test error and 99.5\% sparsity.}
  \label{fig:lenet5}
\end{minipage}
\end{figure}

Figure \ref{fig:lcn_weights_vis} shows a typical LCN filter learned with APT, consisting of 24 $\times$ 24 local filters with unshared weights. The local filters around border regions of the input image appear largely inactive, while those near the center freely developed various shapes with some common structure. We suspect that more training data would not help LCN's learned local filters to converge to the structured parameter tying enforced by CNN (although the error of the LCN could go down), as the discriminatory information in the images is not evenly distributed spatially, and in this case the main appeal of CNN's parameter-tying assumption is higher computational efficiency.

\subsubsection{Sparse APT on LeNets}
Figure \ref{fig:lenet5} visualizes the final weights in LeNet-5's first 20 convolution filters: as can be seen, 11 of them contained zero weights only (thus considered pruned), while the remaining important stroke detectors were quantized. More generally we observed structured sparsity in weights (row/column-wise sparsity for fully connected layers and channel/filter-wise sparsity for conv layers) that result in entire units pruned away, similar to group-sparsity pursued by \citet{wen2016}. \footnote{Our small-scale evaluation using $\ell_1$ alone for compression indicates that this appears to be a general property of lasso penalty; however, training with $\ell_1$ followed by pruning (by setting all parameters below a tuned threshold $\epsilon$) did not achieve as much sparsity as sparse APT for the same accuracy.}

\begin{figure}[t]
    \centering
    \begin{minipage}{0.49\textwidth}
        \centering
        \includegraphics[width=1\textwidth]{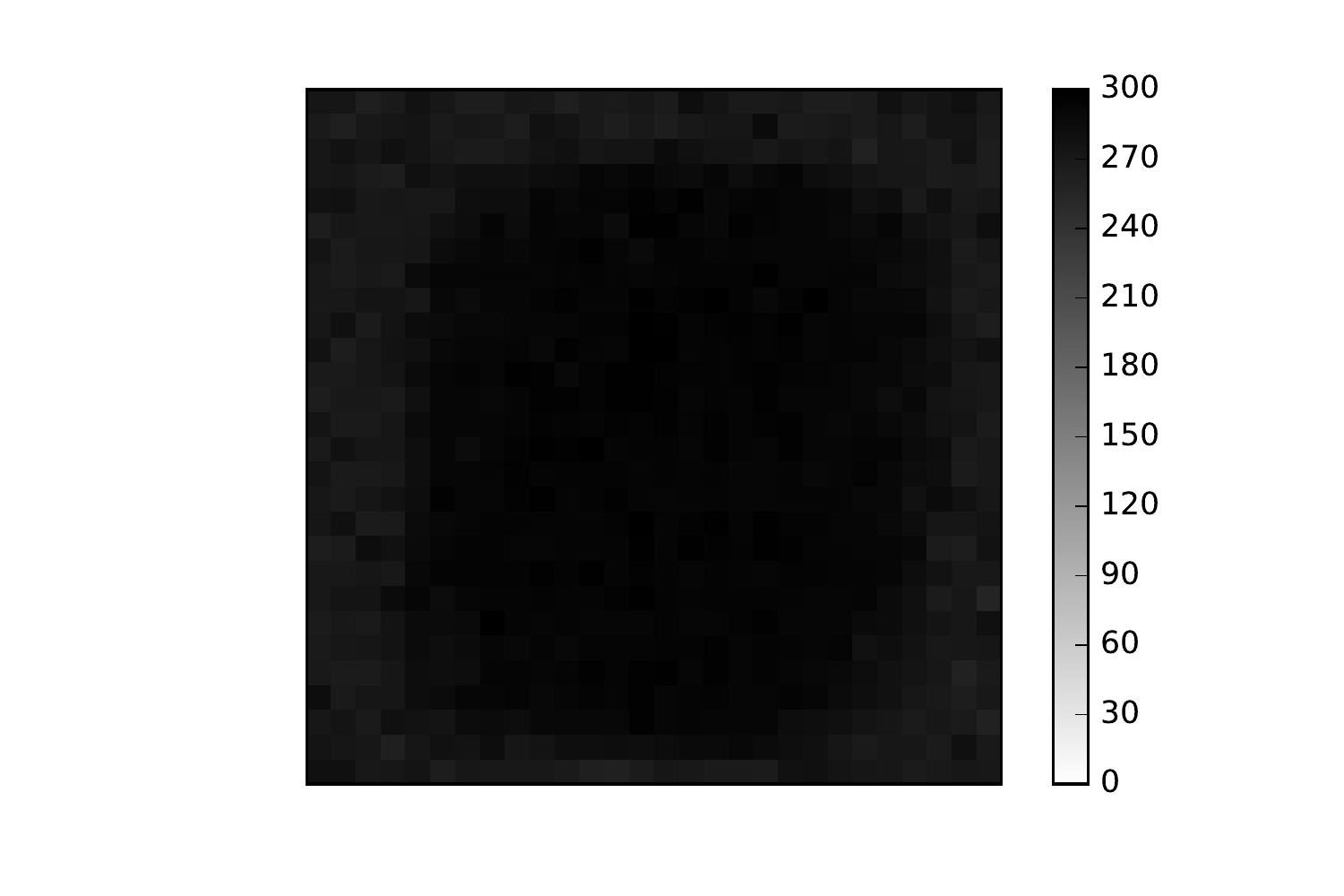} 
    \end{minipage} \hfill
    \begin{minipage}{0.49\textwidth}
        \centering
        \includegraphics[width=1\textwidth]{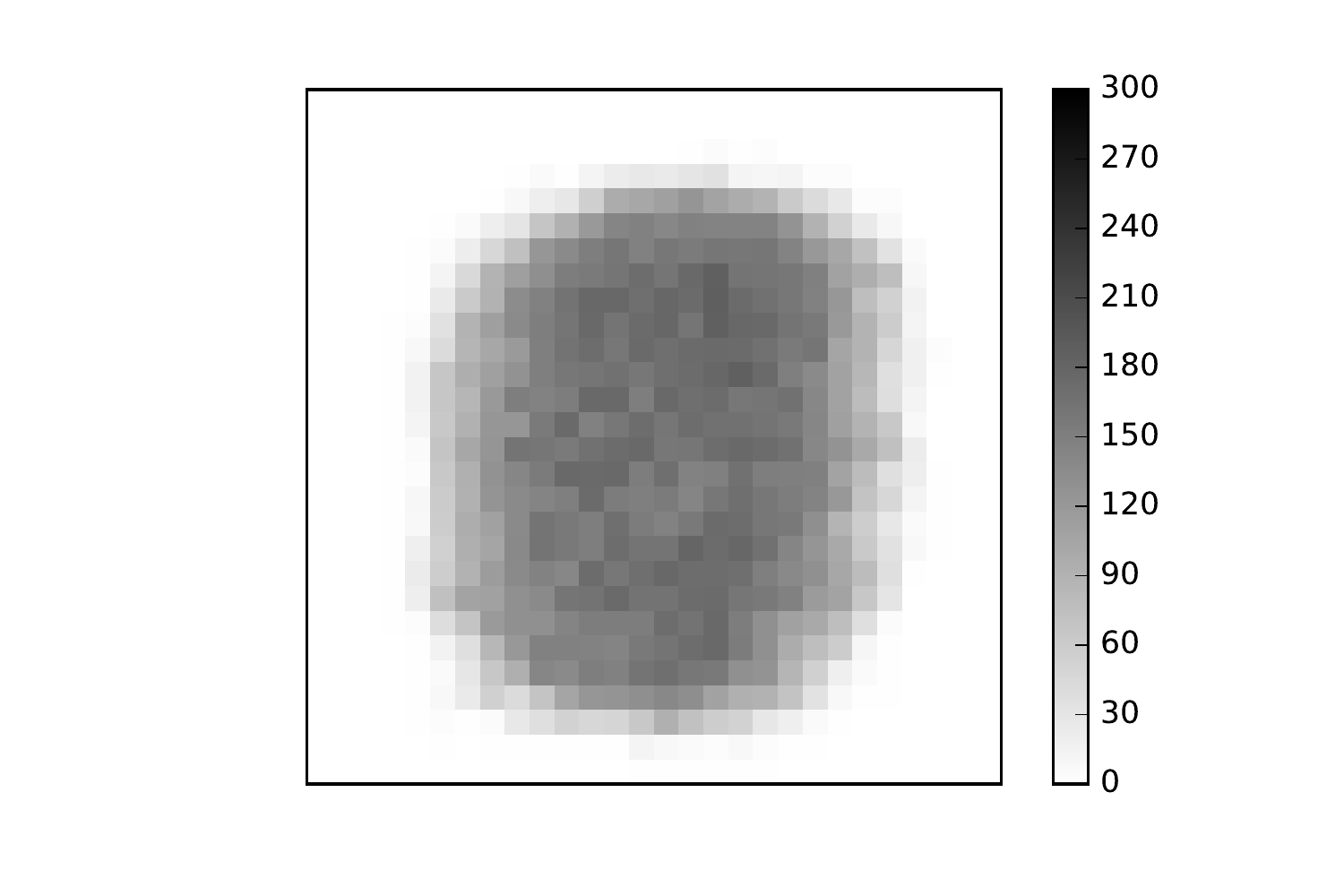} 
    \end{minipage}
        \centering
    \begin{minipage}{0.49\textwidth}
        \centering
        \includegraphics[width=1\textwidth]{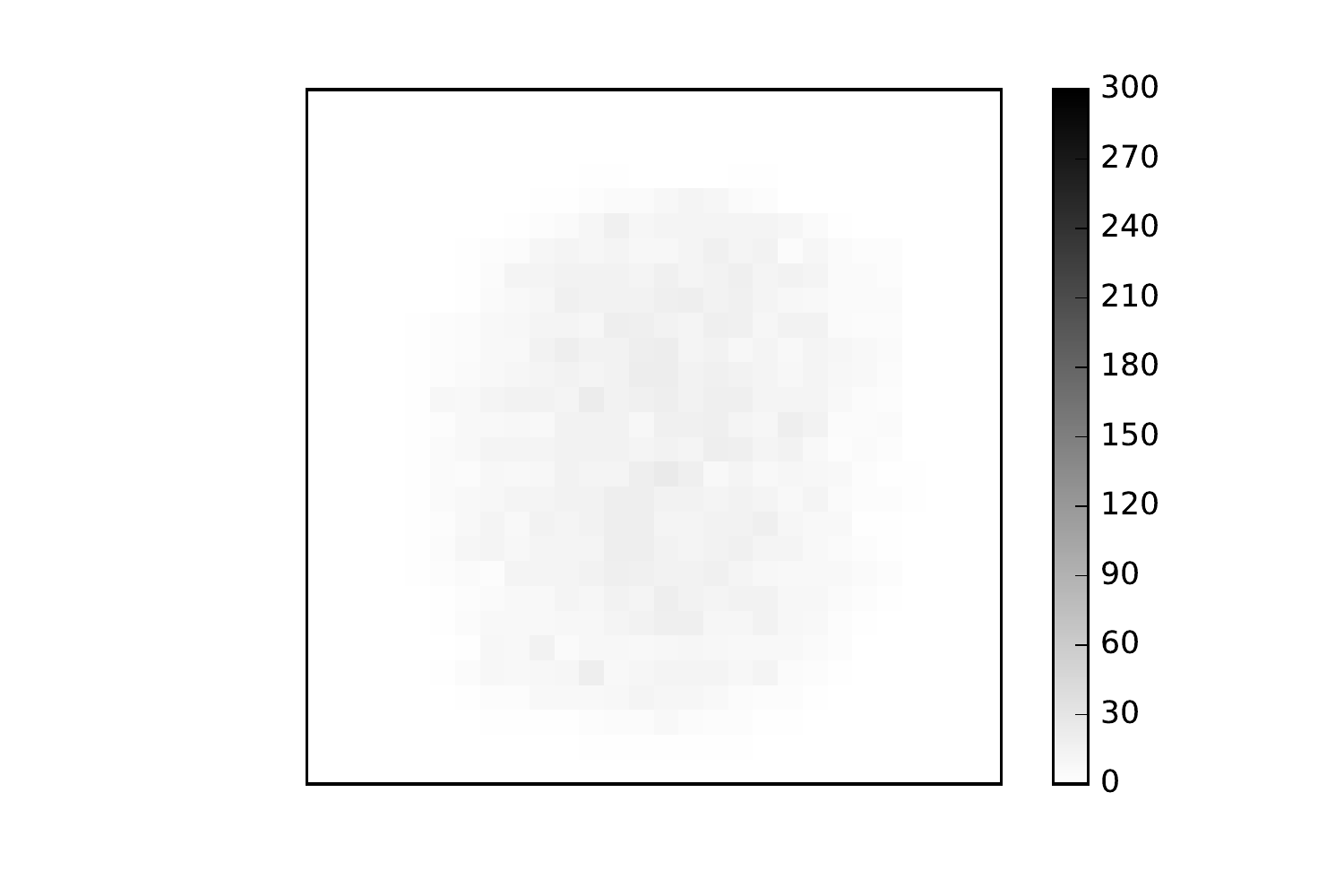} 
    \end{minipage}\hfill
    \caption{Comparing the the number of input units pruned by $\ell_2$, $\ell_1$, and sparse APT, on LeNet-300-100.} \label{fig:lenet300100_input}
\end{figure}

\begin{figure*}[hb]
\centering
\subcaptionbox{$\ell_2$ regularization.}
{\includegraphics[height=.17\paperheight]{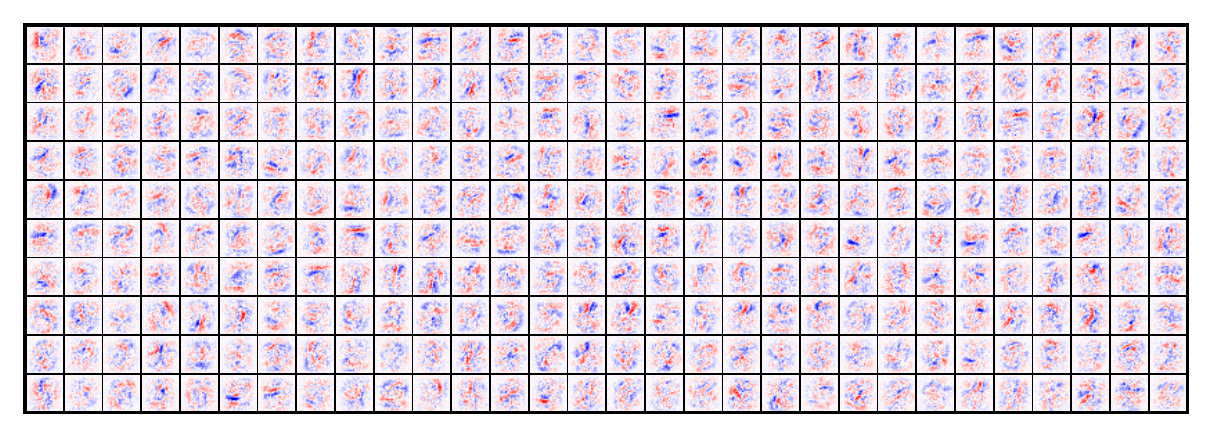}}
\subcaptionbox{APT regularization, with $K=8$ quantization levels.}
{\includegraphics[height=.17\paperheight]{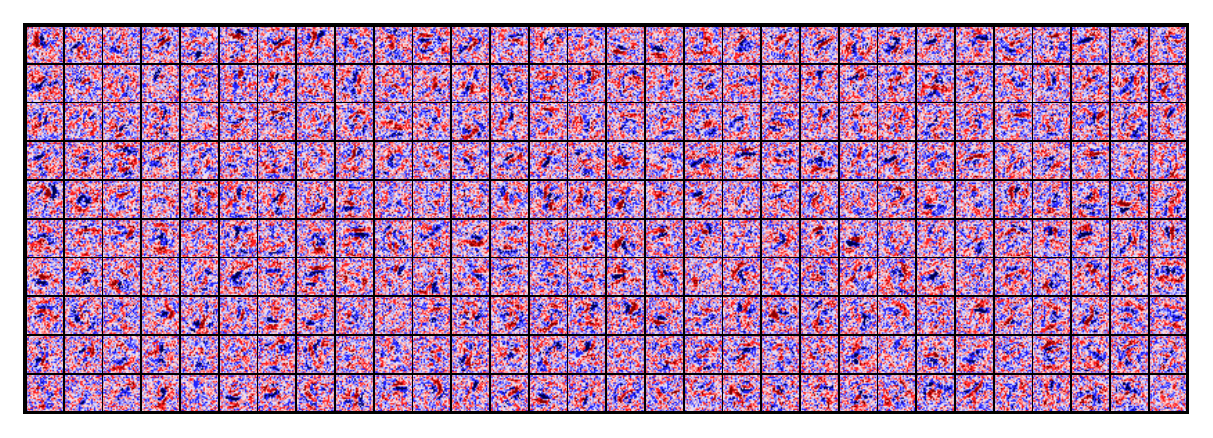}}
\subcaptionbox{$\ell_1$ regularization.}
{\includegraphics[height=.17\paperheight]{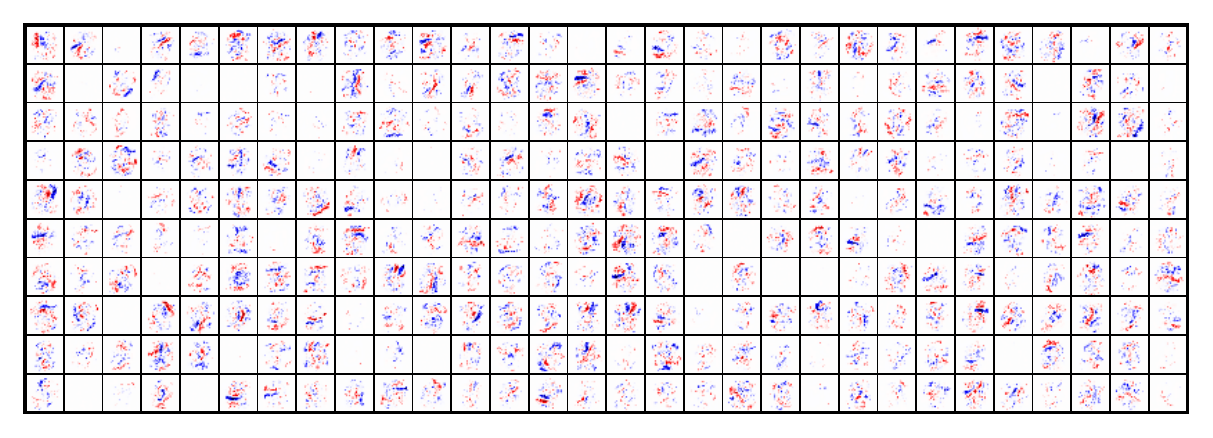}}
\subcaptionbox{Sparse APT regularization, with 16 non-zero quantization levels.}
{\includegraphics[height=.17\paperheight]{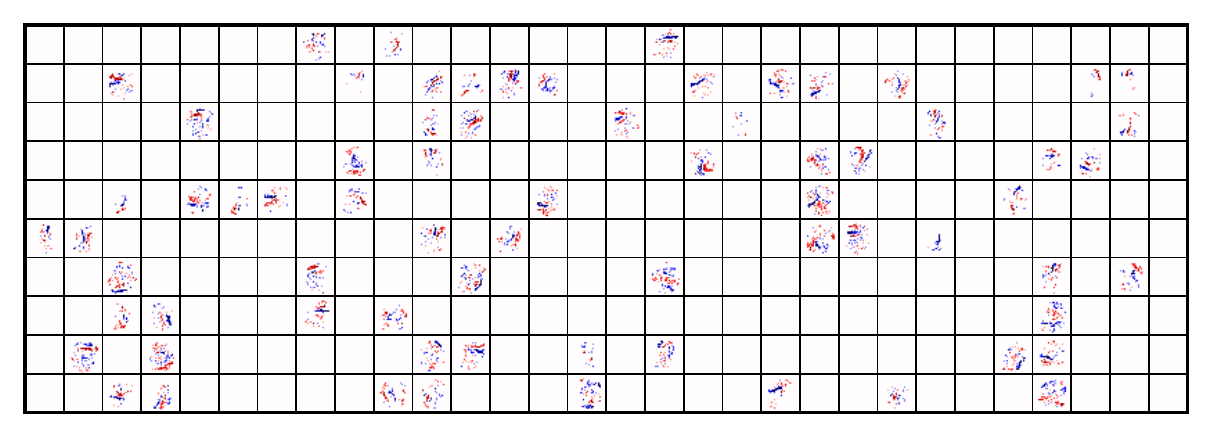}}
\caption{First layer weight matrix of LeNet-300-100 learned with $\ell_2$, APT, $\ell_1$, and sparse APT. } 
\label{fig:lenet300100_layer1}
\end{figure*}

Figure \ref{fig:lenet300100_input} and \ref{fig:lenet300100_layer1} visualize the first layer weights ($300 \times 784$ matrix) of LeNet-300-100 learned with $\ell_2$, $\ell_1$, and (sparse) APT ($K=17$, as reported in table \ref{tab:cr_results}), all starting from the same random initialization and resulting in similar error rates (between 1.8\% and 1.9\%).

Figure \ref{fig:lenet300100_input} plots the count of non-zero outgoing connections from each of the 784 input units (shaped as $28 \times 28$ matrix), to the next layer's 300 hidden units. An input unit is considered pruned if all of its outgoing weights are zero; this corresponds to a column of zeros in the weight matrix. Here, sparse APT prunes away 403 of the 784 input units, giving a column-sparsity of 48.6\%. 

The situation of plain APT is similar to $\ell_2$ and is not shown. In the solutions learned with $\ell_2$ and $\ell_1$, we mark weights with magnitude less than $10^{-3}$ as zero for illustration, since $\ell_2$ and $\ell_1$ did not result in exactly zero weights. 

Figure \ref{fig:lenet300100_layer1} depicts the first layer weight matrix of LeNet-300-100; each of the 784 input connections to the next layer unit are reshaped as a $28 \times 28$ cell. All colors are on an absolute scale from -0.3 to 0.3 centered at 0; thus a white cell indicates a hidden unit has been disconnected from input and degenerated into a bias for the next layer, corresponding to a sparse row in the weight matrix. Sparse APT results in 76.3\% row-sparsity in this case.

\newpage

\end{document}